\documentclass[review, sort&compress]{elsarticle}
\usepackage[breaklinks]{hyperref}
\usepackage{times}
\usepackage{latexsym}

\usepackage{geometry}

\usepackage{subfiles}
\usepackage{multirow}
\usepackage{array}
\usepackage{hyphenat}
\usepackage{subcaption}

\usepackage{graphicx}
\usepackage{adjustbox}   
\usepackage{enumitem}
\usepackage{url}

\usepackage{bm}
\usepackage{booktabs}
\usepackage{float}

\usepackage{soul}
\usepackage{color}

\usepackage[ruled,vlined]{algorithm2e}
\SetAlFnt{\small}
\usepackage{mathtools}
\usepackage{amsmath}
\usepackage{wrapfig}

\newcommand{\wrapvspace}{-0.10in}
\usepackage{tabu}
\usepackage{ctable}

\usepackage{placeins}

\makeatletter
\newcommand{\thickhline}{%
    \noalign {\ifnum 0=`}\fi \hrule height 1pt
    \futurelet \reserved@a \@xhline
}
\newcolumntype{"}{@{\hskip\tabcolsep\vrule width 1pt\hskip\tabcolsep}}
\makeatother

\usepackage{setspace}
\usepackage{lineno}
\modulolinenumbers[5]

\journal{arXiv}

\begin{document}

\begin{frontmatter}

\title{Annotating Social Determinants of Health Using Active Learning, and Characterizing Determinants Using Neural Event Extraction}

\author[1]{Kevin Lybarger}\corref{cor1}
\ead{lybarger@uw.edu}

\author[2]{Mari Ostendorf}
\author[1]{Meliha Yetisgen}

\cortext[cor1]{Corresponding author}

\address[1]{Biomedical \& Health Informatics, University of Washington, Box 358047 Seattle, WA 98109, USA}

\address[2]{Department of Electrical \& Computer Engineering, University of Washington, Campus Box 352500 185, Seattle, WA 98195-2500, USA}

\begin{abstract}

Social determinants of health (SDOH) affect health outcomes, and knowledge of SDOH can inform clinical decision-making. Automatically extracting SDOH information from  clinical text requires data-driven information extraction models trained on annotated corpora that are heterogeneous and frequently include critical SDOH. This work presents a new corpus with SDOH annotations, a novel active learning framework, and the first extraction results on the new corpus. The Social History Annotation Corpus (SHAC) includes 4,480 social history sections with detailed annotation for 12 SDOH characterizing the status, extent, and temporal information of 18K distinct events. We introduce a novel active learning framework that selects samples for annotation using a surrogate text classification task as a proxy for a more complex event extraction task. The active learning framework successfully increases the frequency of health risk factors and improves automatic extraction of these events over undirected annotation. An event extraction model trained on SHAC achieves high extraction performance for substance use status (0.82-0.93 F1), employment status (0.81-0.86 F1), and living status type (0.81-0.93 F1) on data from three institutions. 

\end{abstract}


\end{frontmatter}


\section{Introduction}

US life expectancy is decreasing \citep{murphy2018mortality}, even as medical care advances. Decreasing life expectancy may be partly attributable to deteriorating social determinants of health (SDOH) \citep{daniel2018addressing, himmelstein2018determined}. For example, substance abuse (including alcohol, drug, and tobacco use) is increasingly recognized as a key factor for morbidity and mortality \citep{centers2005annual, world2019global, degenhardt2012extent}. More Americans are living alone, leading to increased social isolation and negative health outcomes \citep{cacioppo2003social}. Employment and occupation impact income, societal status, hazards encountered, and health \citep{clougherty2010work}. Understanding SDOH, including behaviors influenced by these social factors, can inform clinical decision-making \citep{blizinsky2018leveraging}.

SDOH are characterized in the Electronic Health Record through structured data and unstructured clinical text; however, clinical text captures detailed descriptions of these determinants, beyond the representation in structured data. This text-encoded information must be automatically extracted for secondary use applications, like large-scale retrospective studies and clinical decision support systems. The automatically extracted data can augment the available structured data to create a more comprehensive patient representation in these downstream applications \citep{demner2009can, jensen2012mining}.

Leveraging the social history information in clinical text requires high-quality annotated data to create machine learning-based information extraction models. This work presents a new annotated clinical corpus, referred to as  Social History Annotation Corpus (SHAC). \textit{SHAC} is comprised of 4,480 social history sections with detailed annotations for 12 critical SDOH. SHAC utilizes clinical notes from MIMIC-III \citep{johnson2016mimic} and an existing data set from the University of Washington (UW) and Harborview Medical Centers. It includes event-based annotations for more than 55K annotated spans and 18K distinct events across four note types. 

Hand annotation of detailed SDOH information in clinical notes is costly, and many critical SDOH are infrequent. To address these budget and data sparsity limitations, the corpus development used active learning to select samples for annotation. Because extracting the event-based SDOH phenomena is a complex sequence labeling task, standard active learning methods are not practical. This work introduces a novel active learning framework that uses a simplified surrogate task for assessing sample informativeness. Our experiments show that this method increases the diversity and richness of the annotations and improves extraction performance for a variety of event types. The largest performance gains achieved by the active learning framework are associated with infrequent, but extremely important risk factors, like drug use, homelessness, and unemployment.

With the annotated SHAC corpus, we provide a baseline neural event extractor and present the first reported extraction results on SHAC for the most frequently annotated SDOH: substance use, employment, and living status. The event extraction model identifies substance use,  employment, and living status events at 0.89-0.98 F1 and characterizes the status of these determinants with 0.81-0.96 F1. The annotation guidelines and source code will be made available online\footnote{\url{https://github.com/uw-bionlp}}.

\section{Related work}

\subsection{SDOH Corpora}

Multiple corpora with note-level SDOH annotations have been developed. For example, the i2b2 NLP Smoking Challenge introduced a publicly available corpus where tobacco use status is labeled at the note-level \citep{uzuner2008identifying}. \citet{ gehrmann2018comparing} annotated MIMIC-III discharge summaries with note-level phenotype labels, including substance abuse and obesity. \citet{feller2018towards} annotated 38 different SDOH at the note-level. Annotated corpora with more detailed SDOH annotations describing status, extent, temporal information, and other characteristics also exist. For example, \citet{ wang2015automated} introduced a corpus with detailed substance use annotations for 691 clinical notes, and \citet{Yetisgen_2017_substance} created detailed annotations for 13 SDOH in a publicly available corpus of 364 notes. Both \citet{ wang2015automated} and \citet{Yetisgen_2017_substance} utilized deidentified notes from the MTSamples website\footnote{MTSamples website: \url{http://www.mtsamples.com/}} that were created by human transcriptionists. 

To achieve high SDOH extraction performance that generalizes across clinicians, institutions, and specialties, annotated corpora must be sufficiently large and diverse. Unfortunately, existing publicly available corpora with SDOH annotations are lacking in either annotation detail, size, and/or heterogeneity. SHAC provides a relatively large corpus with high quality, detailed SDOH annotations. SHAC is heterogeneous in that it includes clinical notes from multiple institutions and note types, and in the use of active selection to encourage a richer representation of SDOH events.

\subsection{Active Learning}
In annotation projects, the available unlabeled data is often significantly larger than the annotation budget. Randomly selecting samples for annotation is suboptimal from a model learning perspective, as samples vary in their usefulness, particularly when the phenomena of interest may be infrequent. Active learning identifies samples for annotation that maximize model learning \citep{cohn1994improving, cohn1996active}. Samples are selected using a query function that scores sample informativeness, representativeness, and/or diversity \citep{shen2004multi, yang2015multi, du2017exploring}. Informativeness describes the potential for a sample to reduce classification uncertainty. The literature varies in the usage of the terms ``representativeness'' and ``diversity.'' Here, representativeness describes the degree to which a sample describes the structure of the data, and diversity characterizes the variation in the samples selected. 

Active learning is well-established for classification tasks, where a single label is predicted for each sample. Multiple studies have applied active learning to text classification tasks, where a sample is a sentence or a document. Sample informativeness is derived from classification uncertainty scores, such as maximizing entropy \citep{Wu2013graph} or minimizing a support vector machine margin \citep{TongKoller2002,PARK2015efficient}. \citet{du2017exploring} assesses diversity based on classifier posterior distributions, and \citet{Wu2013graph} assesses diversity and representativeness based on sample similarity within the observation space.

Approaches for applying active learning to sequence tagging problems are also well-established \citep{chen2015a_study, chen2017an_active, Kholghi2017clnical, li2019efficient, Gao2019recognizing, Shelmanov2019active}. Although predictions are made at the token-level, sample selection is typically performed at the sentence or document-level. Representativeness and/or diversity are often assessed by calculating sentence similarity metrics in the observation space \citep{chen2015a_study, chen2017an_active, Kholghi2017clnical, Gao2019recognizing}. Sequence-level uncertainty scores are calculated by various measures, like normalized prediction sequence likelihood and minimum token-level confidence.
In the clinical and biomedical domain, uncertainty scores are generated with conditional random field (CRF) models \citep{chen2015a_study, chen2017an_active, Kholghi2017clnical, li2019efficient, Gao2019recognizing} or a neural tagger based on contextualized embeddings from ELMo and BERT \citep{Shelmanov2019active}.

Active learning is less explored in relation and event extraction tasks, where triggers (heads), arguments, and/or relations are annotated. The predictions are more complex, involving labeling and linking spans of text. \citet{maldonado2017active} apply active learning to a clinical relation extraction task, selecting samples using the average entropy of all predicted phenomena as an uncertainty score. More recently, \citet{MALDONADO2019103265active} explores active learning in a medical concept and relation extraction task. In lieu of a heuristic query function, an optimal selection strategy is learned from data with strong and weakly supervised labels, including 1,000 electroencephalogram (EEG) reports with automatic annotations generated by existing extraction models. 

SHAC is annotated using an event-based structure, where SDOH are characterized through multiple argument types. These argument types are not equally important for secondary use applications, and the entropy of different determinant-argument combinations may differ significantly. Without sufficient annotated data to learn an optimal selection strategy, we use a simplified text classification task as a surrogate for assessing sample uncertainty, to prevent under sampling the critical phenomena. We hypothesized that the surrogate task would improve extraction performance in the more complex event extraction task and validated the hypothesis with experiments on SHAC data.


\section{Materials}
\label{materials}

\subsection{Data}
This work utilized two clinical data sets without SDOH annotations: \textit{MIMIC-III} and \textit{UW Dataset}. \textit{MIMIC-III} (referred to here as \textit{MIMIC}) is a publicly available, deidentified health database for over 40K critical care patients at Beth Israel Deaconess Medical Center from 2001-2012 \citep{johnson2016mimic}. MIMIC contains clinical notes, diagnosis codes, and other data. This work utilized 60K MIMIC discharge summaries. The \textit{UW Dataset} is an existing clinical data set from the UW and Harborview Medical Centers generated between 2008-2019. This work utilized 83K emergency department, 22K admit, 8K progress, and 5K discharge summary notes from UW Dataset. An existing corpus with SDOH annotations created by Yetisgen and Vanderwende, \textit{YVnotes}, was used for model training during active learning \citep{Yetisgen_2017_substance}.

\subsection{Annotation Scheme}
We created detailed annotation guidelines for 12 SDOH (referred to here as \textit{event types}), including substance use (alcohol, drug, and tobacco), physical activity, employment, insurance, living status, sexual orientation, gender identity, country of origin, race, and environmental exposure. Each \textit{event} is a characterization of a specific SDOH instance and includes a trigger (head) and all associated arguments (attributes). These \textit{events} capture changes to the status, extent, and temporality of SDOH in the patient timeline. Each event type is annotated across multiple dimensions. Table \ref{annotated_phenomena_partial} summarizes the annotation of the most frequent SHAC event types: substance use, employment, and living status. Table \ref{annotated_phenomena_all} in the Appendix contains a summary of all annotated event types. 
\begin{table}[!ht]
    \small
    \centering
    
\begin{tabular}{|p{0.9in}|p{0.8in}|p{2.0in}|p{1.45in}|}
\hline
\textbf{Event type, $e$}                                         & \textbf{Argument type, $a$} & \textbf{Argument subtypes, $\bm{y_l}$}                                                  & \textbf{Span examples}                                            \\ \hline
\multirow{8}{0.8in}[\baselineskip]{Substance use (Alcohol, Drug, \& Tobacco)} & Status\textsuperscript{*}            & \{none, current, past\}                           & ``denies," ``smokes"                             \\ \cline{2-4} 
                                                            & Duration          & --                                                                  & ``for the past 8 years"                                            \\ \cline{2-4} 
                                                            & History           & --                                                                  & ``seven years ago"                                                 \\ \cline{2-4} 
                                                            & Type              & --                                                                  & ``beer," ``cocaine"                                         \\ \cline{2-4} 
                                                            & Amount            & --                                                                  & ``2 packs," ``3 drinks"                                           \\ \cline{2-4} 
                                                            & Frequency         & --                                                                  & ``daily," ``monthly"                                      \\ \hline 
\multirow{4}{0.8in}[\baselineskip]{Employment}                             & Status\textsuperscript{*}            & \{\nohyphens{employed, unemployed, retired, \newline on disability, student, homemaker}\} & ``works," ``unemployed"               \\ \cline{2-4} 
                                                            & Duration          & --                                                                  & ``for five years"                                         \\ \cline{2-4} 
                                                            & History           & --                                                                  & ``15 years ago"                                                    \\ \cline{2-4} 
                                                            & Type              & --                                                                  & ``nurse," ``office work"                                 \\ \hline
\multirow{4}{0.8in}[\baselineskip]{Living status}                          & Status\textsuperscript{*}            & \{current, past, future\}                            & ``lives," ``lived"                                         \\ \cline{2-4} 
                                                            & Type\textsuperscript{*}              & \{\nohyphens{alone, with family, with others, homeless}\}     & ``with husband," ``alone"                        \\ \cline{2-4} 
                                                            & Duration          & --                                                                  & ``for the past 6 months"                                         \\ \cline{2-4} 
                                                            & History           & --                                                                  & ``until a month ago"                                               \\ \hline

\end{tabular}

    \caption{Annotation guideline summary for the most frequent event types. *indicates the argument is required.}
    \label{annotated_phenomena_partial}
\end{table}

\begin{wrapfigure}{R}{0.605\textwidth}
    \frame{\includegraphics[width=0.6\textwidth]{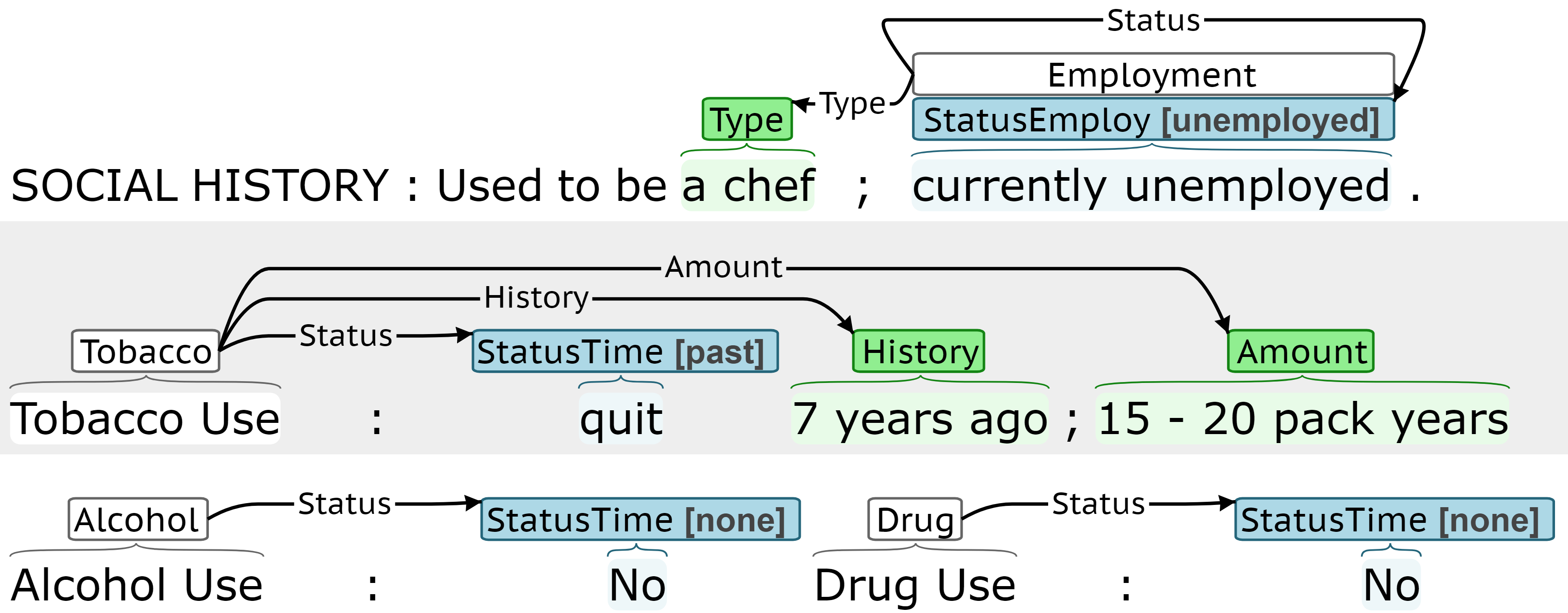}}
    \caption{BRAT annotation example}
    \label{brat_example}
\end{wrapfigure}
SDOH are annotated as events using the BRAT rapid annotation tool \citep{stenetorp2012brat}. Figure \ref{brat_example} is a BRAT annotation example, describing a patient's employment and substance use. The trigger indicates the event type (e.g. \textit{Employment} or \textit{Tobacco}) and arguments describe the event. \textit{Labeled arguments}, like \textit{Status}, include both an annotated span and subtype label. \textit{Span-only arguments}, like \textit{Duration} or \textit{History}, include an annotated span without an additional subtype. 

\subsection{Annotation Cycle}
Social history sections, referred to here as \textit{samples}, were extracted from MIMIC and the UW Dataset, using pattern matching to identify section headings (alphanumeric, forward slash, backslash, ampersand, or white space characters followed by a colon). SHAC includes \textit{train}, \textit{development}, and \textit{test} sets. Samples for the train set were randomly and actively selected. Training samples were randomly selected for initial model training in active learning, then the initial model Was used in actively selecting samples to bias the training set towards diverse samples that frequently contain the phenomena of interest. All development and test samples were randomly selected to approximate the true distribution of the SDOH in the corpora used. Samples were annotated by four medical students through 12 rounds of annotation (8 randomly selected and 4 actively selected). Table \ref{annotation_rounds} in the Appendix describes each round of annotation. The first two rounds were randomly sampled and double-annotated, to assess inter-annotator agreement. After the initial annotation round, the annotation guidelines were revised, and the initial annotations were updated.

\subsection{Evaluation and Annotation Scoring}
\label{annotator_agreement_scoring}
\begin{figure}[H]
    \centering
    \frame{\includegraphics[width=0.9\textwidth]{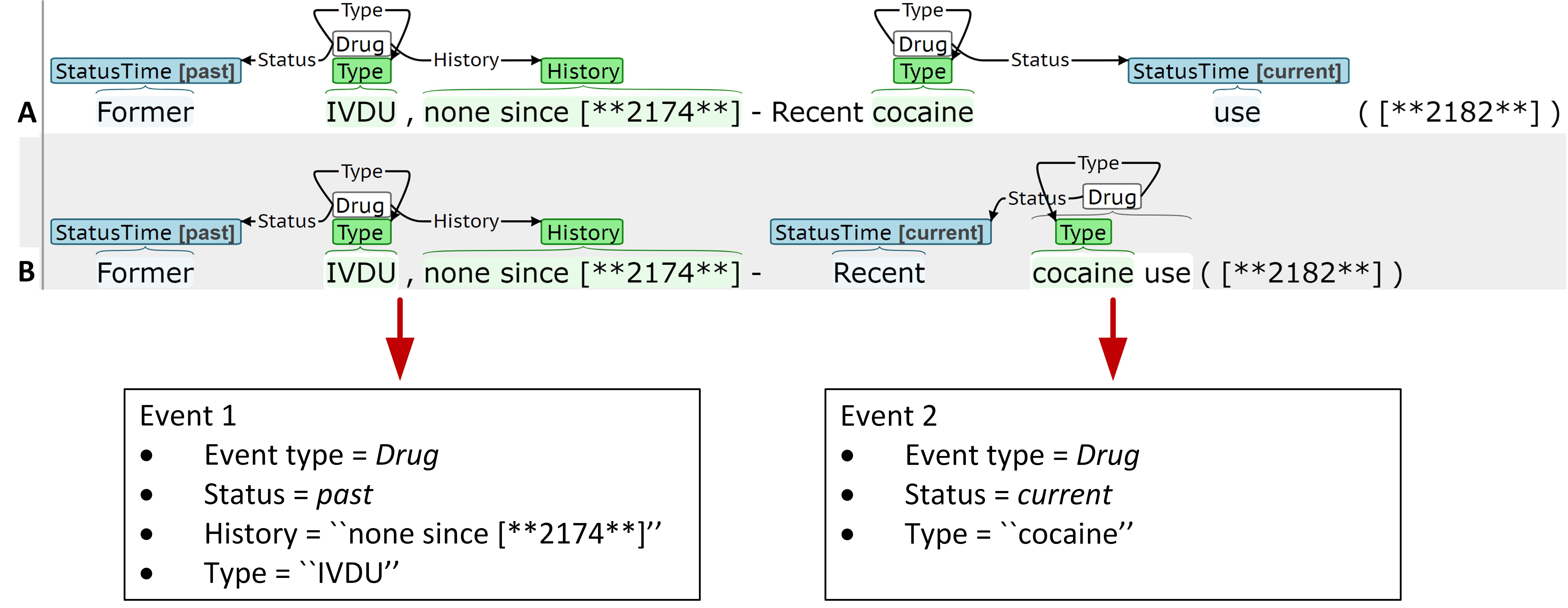}}
    \caption{Annotation examples describing event extraction as a slot filling task}
    \label{annotation_comparison}
\end{figure}

We treat event annotation and extraction as a slot filling task, as this is most relevant to secondary use applications. As such, there can be multiple equivalent span annotations. Figure \ref{annotation_comparison} presents the same sentence annotated by two annotators (labeled \textit{A} and \textit{B}), along with the populated slots.
\noindent
Both annotators labeled two \textit{Drug} events: \textit{Event 1} and \textit{Event 2}. Event 1 describes past intravenous drug use (IVDU), and Event 2 describes current cocaine use. Event 1 is annotated identically by both annotators. However, there are differences in the annotation spans of Event 2, specifically for the \textit{Trigger} (``cocaine" versus ``cocaine use") and \textit{Status} (``use" vs. ``Recent"). From a slot perspective, the annotations for Event 2 are equivalent. Thus, scoring of automatic detection and annotator agreement is based on relaxed span match criteria, as described below. Trigger and argument performance is evaluated using precision (P), recall (R), and F1, micro averaged over the event types, argument types, and/or argument subtypes.

\textbf{Trigger:} Triggers, $T_i$, are represented by a pair (event type, $e_i$; token indices, $x_i$). For \textit{Event 2} in Figure \ref{annotation_comparison}, $T_{A,2}=(e_{A,2}=Drug; x_{A,2}=[8])$ and $T_{B,2}=(e_{B,2}=Drug; x_{B,2}=[8,9])$. Triggers of the same event type, $e$, are aligned by minimizing the distance between span centers computed from the token indices. Trigger equivalence is defined as 
\begin{equation}
T_i \equiv T_j \mbox{ if } (e_i \equiv e_j) \land (T_i \mbox{ aligned with } T_j).    
\end{equation}
Although there are two drug events in the Figure \ref{annotation_comparison} example, $T_{A,2}$ aligns with $T_{B,2}$ because of the overlapping spans.

\textbf{Argument:} Events are aligned based on trigger equivalence, and the arguments of aligned events are compared using different criteria for \textit{labeled arguments} and \textit{span-only arguments}. Labeled arguments, $L_i$, are represented as a triple (argument type, $a_i$; token indices, $x_i$; subtype, $l_i$). For \textit{Event 2} in Figure \ref{annotation_comparison}, $L_{A,2}=(a_{A,2}=Status; x_{A,2}=[9], l_{A,2}=current)$ and $L_{B,2}=(a_{B,2}=Status; x_{B,2}=[7], l_{B,2}=current)$. For labeled arguments, the argument type, $a$, and subtype, $l$, capture the salient information and equivalence is defined as 

\begin{equation}
L_i \equiv L_j \mbox{ if } (T_i \equiv T_j) \land (a_i \equiv a_j) \land (l_i \equiv l_j).
\end{equation}

Span-only arguments, $S_i$, are represented as a pair (argument type, $a_i$; token indices, $x_i$). For \textit{Event 2} in Figure \ref{annotation_comparison}, $S_{A,3}=(a_{A,3}=Type; x_{A,3}=[7])$ corresponds to ``cocaine.'' Span-only arguments are not easily mapped to a fixed set of classes, and the identified span, $x$, contains the most salient argument information. Span-only arguments with equivalent triggers and argument types, $(T_i \equiv T_j) \land (a_i \equiv a_j)$, are compared at the token-level (rather than the span-level) to allow partial matches. Partial match scoring is used as partial matches can still contain useful information. 

\textbf{Cohen's Kappa:} We evaluate annotator agreement using Cohen’s Kappa, $\kappa$, coefficient, where higher $\kappa$ denotes better annotator agreement \citep{cohen1960coefficient}. Calculating $\kappa$ for the full event structure is not informative, because the probability of random agreement is close to zero. Instead, we calculate $\kappa$ for trigger annotation in the subset of sentences with zero or one trigger for a given event type in either set of annotations, which covers most of the data. We focus on this subset of sentences, because triggers for a given event type are equivalent, if the annotated sentences both include one trigger of that type. We assess annotator agreement on the full event structure using F1 scores.

\subsection{Annotation Statistics}
\begin{figure}[ht]
\centering
\begin{minipage}[b][][b]{.4\textwidth}
    \centering
    \frame{\includegraphics[]{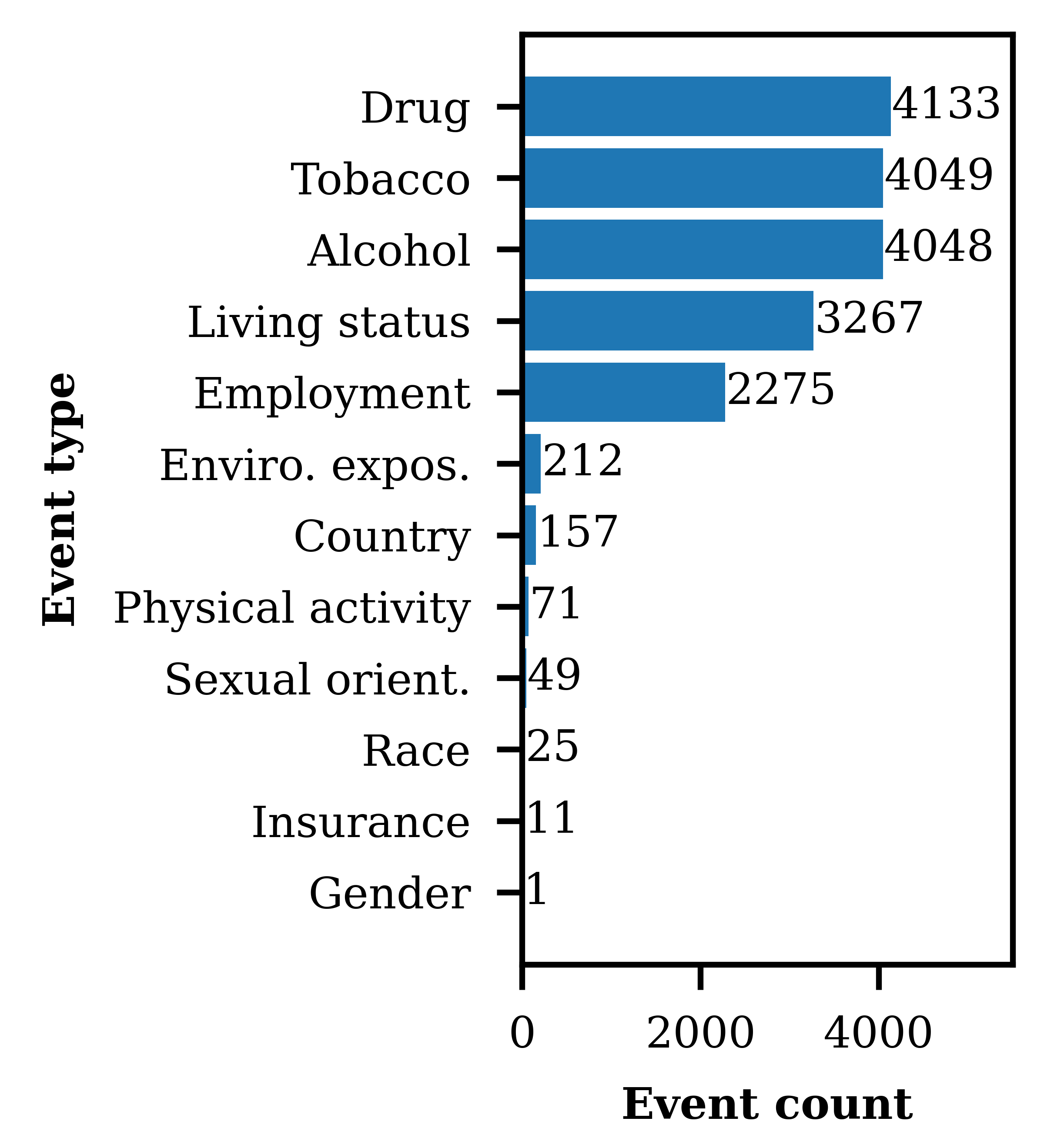}}
    \caption{Event type distribution}
    \label{event_dist}
\end{minipage}%
\begin{minipage}[b][][b]{.1\textwidth}
\hfill
\end{minipage}%
\begin{minipage}[b][][b]{.4\textwidth}
    \centering
    \frame{\includegraphics[]{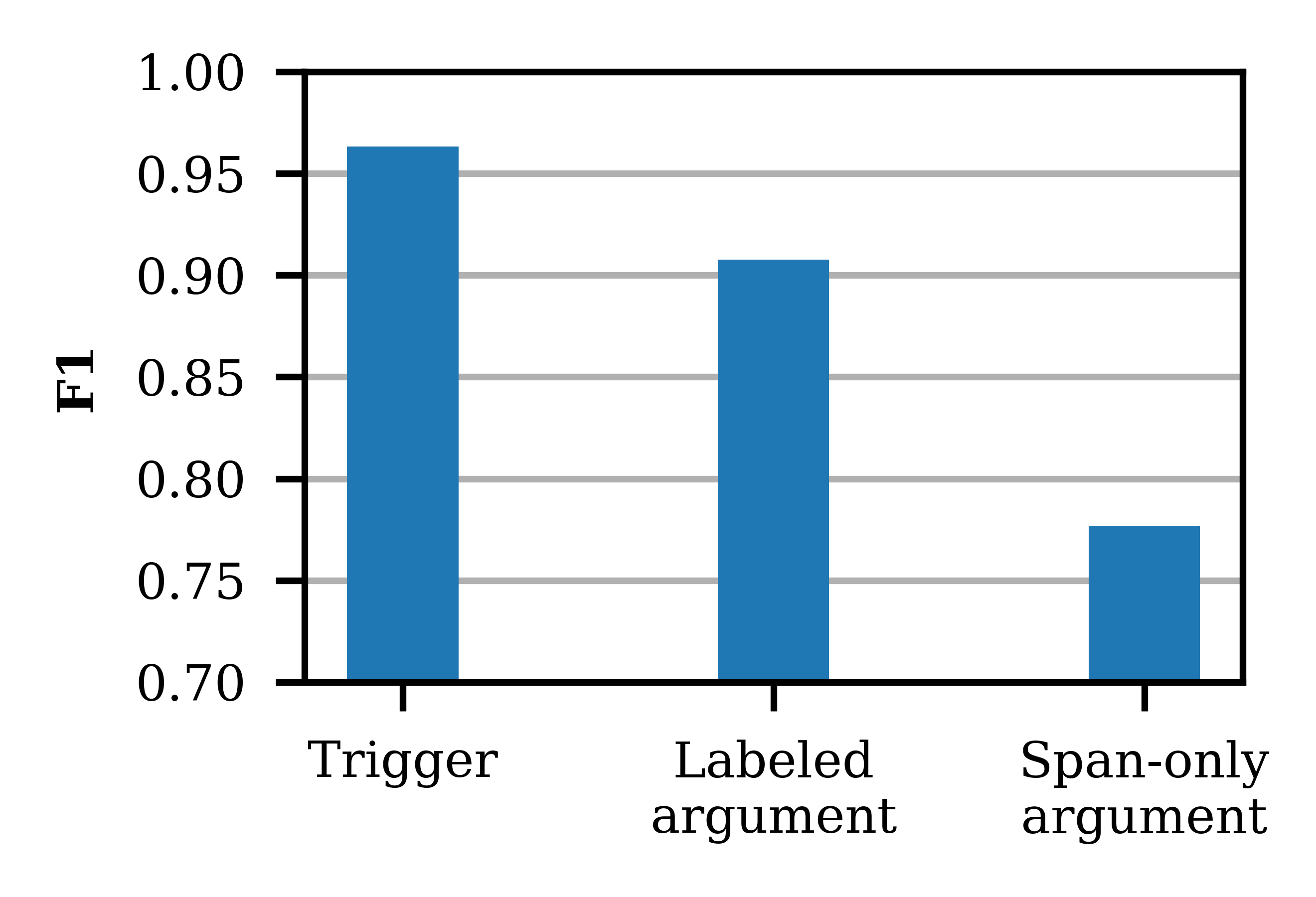}}
    \caption{Annotator agreement for 300 doubly annotated MIMIC samples}
    \label{annotator_agreement}
\end{minipage}
\end{figure}

\begin{wraptable}[]{r}{0.38\textwidth}
    \centering
    \small

\begin{tabular}{|l|c|c|c|}
\hline
\textbf{Source} & \textbf{Train} & \textbf{Dev}    & \textbf{Test}   \\ \hline
MIMIC           & 1,316    & 188 & 376             \\ \hline
UW Dataset      & 1,820    & 260 & 520             \\ \hline
TOTAL           & 3,136    & 448 & 896             \\ \hline
\end{tabular}


    \caption{Corpus composition by source}
    \label{corpus_composition}
\end{wraptable}
SHAC consists of 4,480 annotated social history sections (70\% train, 10\% development, 20\% test). Table \ref{corpus_composition} presents the corpus composition by source. The SHAC training samples are 29\% randomly selected and 71\% actively selected. All development and test data are randomly sampled. Figure \ref{event_dist} presents the event type distribution. The most frequent event types are \textit{Drug}, \textit{Tobacco}, \textit{Alcohol}, \textit{Living status}, and \textit{Employment}, with the remaining event types occurring infrequently. 

Figure \ref{annotator_agreement} presents the annotator agreement for all event types in terms of F1 score for 300 doubly annotated notes from the first two rounds of annotation. For \textit{Alcohol}, \textit{Drug}, \textit{Tobacco}, \textit{Employment}, and \textit{Living status}, trigger $\kappa$ is $0.94-0.97$. For the remaining event types, trigger $\kappa$ is $0.61-0.90$. $\kappa$ is calculated for sentences with 0-1 events for each type ($\ge 99\%$ of all sentences). The trigger agreement is very high, in terms of F1 and $\kappa$, indicating the annotators are consistently identifying and distinguishing between events. The argument agreement is also high for labeled arguments. The somewhat lower agreement for span-only arguments is primarily due to small differences in the start and end token spans (e.g. ``construction worker'' vs. ``construction'').

\section{Active Learning}
\label{active_learning}

This section presents the active learning framework used create SHAC and describes the associated performance gains.

\subsection{Methods}
\begin{wrapfigure}{r}{0.4\textwidth}
    \centering
    \frame{\includegraphics[]{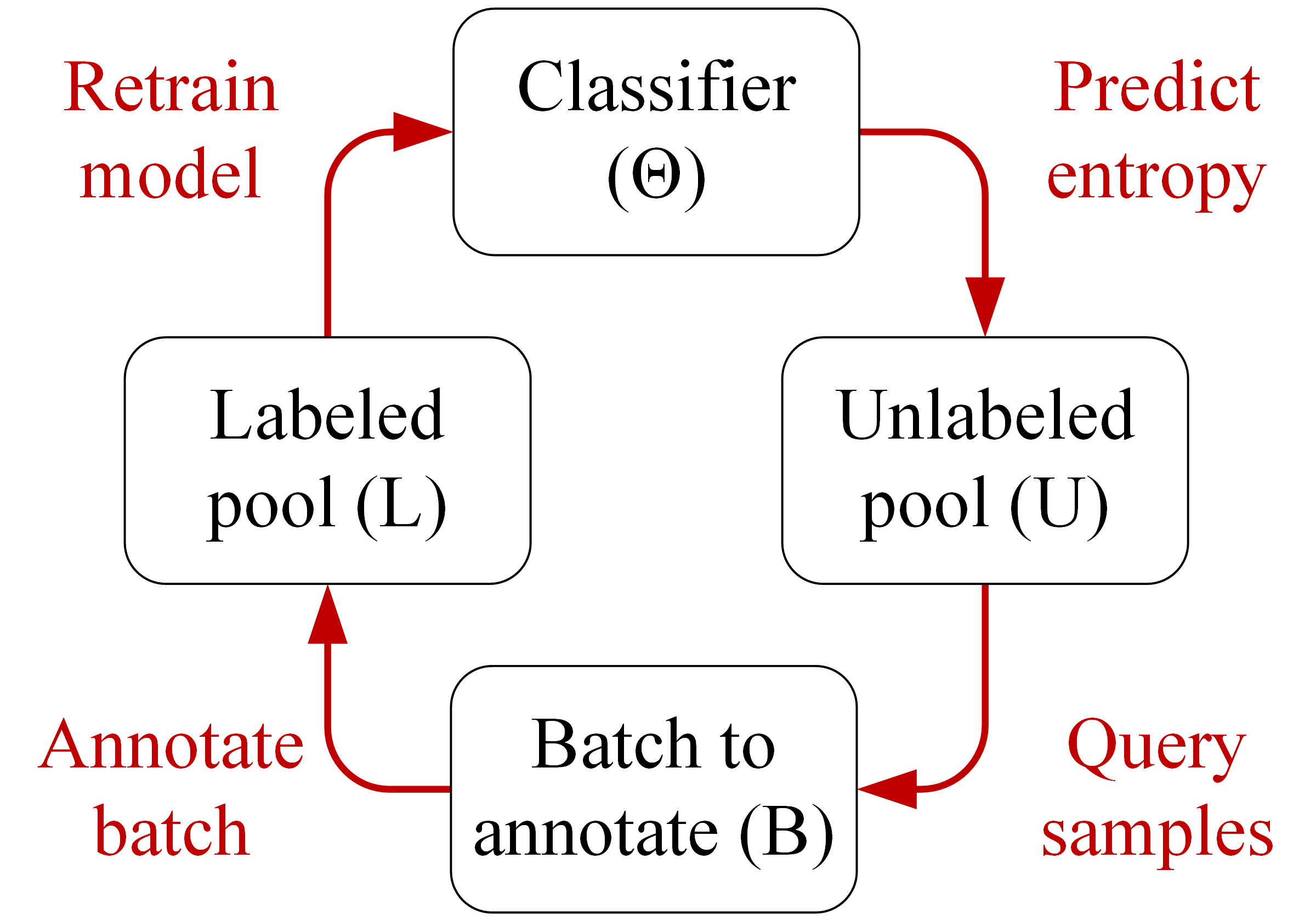}}
    \caption{Active learning annotation cycle}
    \label{active_learning_cycle}
\end{wrapfigure}

A portion of the SHAC training samples were selected using active learning, where a sample is a social history section. Specifically, batch-mode active learning was used to facilitate coordination with human annotators through the cyclical process shown in Figure \ref{active_learning_cycle}. 
\noindent
A batch of samples, $B$, was annotated and added to the labeled pool, $L$. The surrogate classifier was trained on $L$ and then generated uncertainty scores for unlabeled data $U$. Using the uncertainty scores, the query function identified the next batch of samples, $B$. This process was repeated until the annotation objective was met. 
 
Similar to \citet{Wu2013graph}, a query score is designed to combine informativeness and diversity scores of a batch of samples, $B$.  Here, the score has the form: 
\begin{equation} 
Q(B)=\sum_{i \in B}(1-s_i)^\alpha  u(i)
\label{ng_new}
\end{equation}
where $u(i)$ is the uncertainty entropy of sample $i$, $s_i$ is the similarity score of sample $i$ relative to $B$, and $(1 - s_i)$ is the diversity score. $\alpha$ is a weight used to balance the relative importance of the two scores ($\alpha > 0$). The objective is to maximize the batch score, $Q(B)$. We explored different forms for the uncertainty and similarity scores for this multi-label scenario. We implemented a greedy approach to selecting examples, as shown in Algorithm \ref{query_function}. 

\begin{wrapfigure}{r}{0.50\textwidth}
    \centering
    \begin{algorithm}[H]
    \SetAlgoLined
    \KwIn{unlabeled samples $U$, batch size $N$
    }
    \KwOut{batch of samples $B$}
    $B \gets \emptyset$\;
    \While {$\left| B \right| < N$}{
        $k \gets \operatorname*{argmax}_{i \epsilon U} Q(B \cup i)$\;
        $B \gets B \cup \{k\}$\;
        $U \gets U - \{k\}$\;
    }
    \caption{Greedy query function}  
    \label{query_function}
    \end{algorithm}
  
\end{wrapfigure}

\textbf{Diversity:} Sample diversity is assessed in the observation space using two different similarity metrics: \textit{average similarity} and \textit{maximum similarity}, defined as 
\begin{equation*} 
s_{i}^a = \frac{1}{|B|}\sum_{j \in B, j \ne i} a_{j,i}
\qquad s_{i}^m= \max_{j \in B, j \ne i} a_{j,i},
\end{equation*}
respectively,
where $a_{j,i}$ is the cosine similarity of samples $j$ and $i$. 
The maximum similarity approach is a stricter condition that pushes the batch of samples farther apart in the observation space, especially with larger batch sizes. Similar to \citet{lilleberg2015support}, unsupervised vector representations of samples were learned as the TF-IDF weighted averages of pre-trained word embeddings. Word embeddings were created using the word2vec skip-gram model \citep{Mikolov_2013_word2vec} and trained on the entirety of the MIMIC discharge summaries (not just the social history sections). Separate TF-IDF weights were calculated for MIMIC and UW Dataset samples.

\textbf{Uncertainty:}  
Active learning query functions typically assess sample informativeness (uncertainty) using the target classification task. In this work, sample uncertainty was assessed using a simplified surrogate classification task, as a proxy for the more complex event-based annotation scheme. The SHAC annotation scheme includes some arguments (e.g. \textit{Status} for \textit{Alcohol}) that are more predictive of negative health outcomes than others (e.g. \textit{Type} for \textit{Alcohol}), and the prediction uncertainty varies across event types and arguments. To ensure the query function biases selection towards the most salient arguments, each of the five most frequent event types in SHAC were represented using the single argument that is most predictive of negative health outcomes: \textit{Alcohol}-\textit{Status}, \textit{Drug}-\textit{Status}, \textit{Tobacco}-\textit{Status}, \textit{Employment}-\textit{Type}, and \textit{Living status}-\textit{Status}. To cover samples with multiple events of the same type (e.g. both previous and current tobacco use described), an additional class, ``multiple,'' is added to the argument subtypes, $y_l$, in Table \ref{annotated_phenomena_partial}, $y_{c}=\{y_l \cup ``multiple"\}$.

\begin{wrapfigure}{r}{0.35\textwidth}
    \centering
    \frame{\includegraphics[]{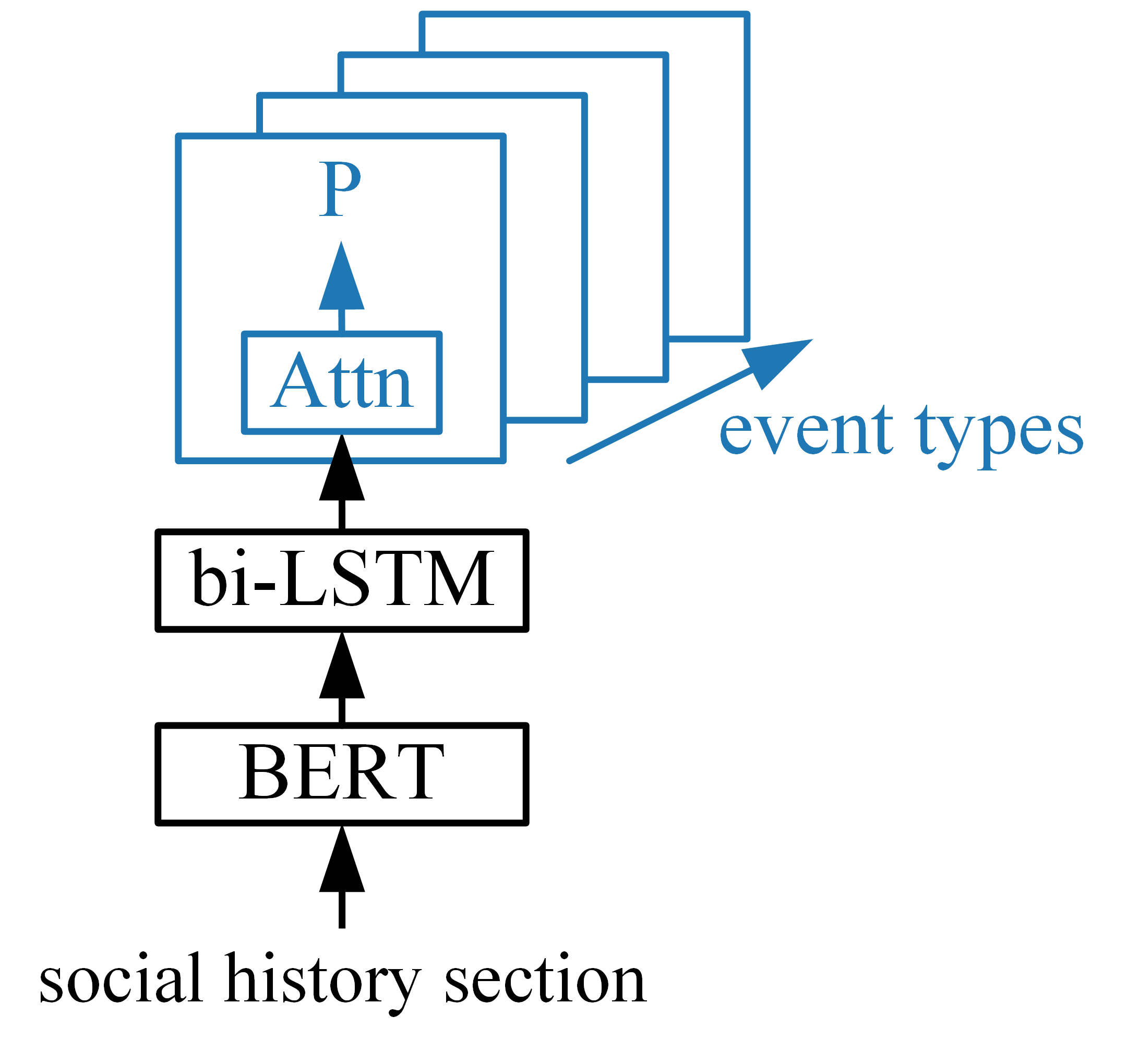}}
    \caption{Surrogate Classifier used to assess sample uncertainty in active learning}
    \label{event_detect}
\end{wrapfigure}

The text classification model, \textit{Surrogate Classifier} in Figure \ref{event_detect}, was used to assess sample uncertainty. The Surrogate Classifier operates on a sample, as a single sequence of $n$ tokens without line breaks. The input social history section is mapped to contextualized word embeddings using \textit{Bio+Discharge Summary BERT} \citep{alsentzer-etal-2019-publicly}, a version of \textit{BERT} \citep{devlin2019bert} trained on clinical text from MIMIC. The BERT output feeds into a bidirectional long short-term memory (bi-LSTM) layer, the output of which feeds into event-specific output layers. Separate self-attention (Attn) output layers for each event type make sample-level predictions. Details of the Surrogate Classifier are similar to the shared and event-argument layers of the full event detection system described in the next section. The Surrogate Classifier generates a set of five multi-class predictions for each sample, one for each event type.

We explored two approaches to characterizing sample uncertainty: i)~the sum of the five event entropy values, similar to previous work \citep{Yang2009effective, wu2014multi, maldonado2017active, Reyes2018evolutionary}, and ii)~entropy for an individual event type, iterating over all types (referred to as ``loop''). As a ``loop'' example, \textit{Alcohol}-\textit{Status} entropy is used for sample 1, \textit{Drug}-\textit{Status} entropy is used for sample 2, and so forth, starting over with \textit{Alcohol}-\textit{Status} entropy for sample 6. The second method was motivated by the concern that summing the entropy values (referred to as ``sum'') could overly bias the selection process in favor of high-entropy event types, reducing the diversity of event types. 

\subsection{Experiments \& Results}
\textbf{Query strategy selection:} Due to limitations in the annotation budget, the query strategy was determined early in the annotation effort. We used the first 700 annotated samples, $L_{Q}$, which consists of random MIMIC samples. $L_{Q}$ was partitioned into $L_{Q}^T\coloneqq \{\mbox{620 train samples}\}$ and  $L_{Q}^D\coloneqq \{\mbox{80 development samples}\}$. For random sampling and each active sampling configuration, 10 runs were performed: 
\begin{enumerate}[label=(\roman*),nolistsep]
    \item $L_{T1} \gets 100 \mbox{ samples from } L_{Q}^T$. Train model, $M_1$, on $L_{T1}$.
    \item $L_{T2} \gets 100 \mbox{ samples from } \{L_{Q}^T - L_{T1}\}$ (random or active). Train model, $M_2$, on $\{L_{T1} \cup L_{T2}\}$.
    \item Evaluate the performance of $M_2$ on $L_{Q}^D$
\end{enumerate}
Active sampling experimentation included different uncertainty types (``loop'' vs.\ ``sum''), similarity types (``average'' vs.\ ``maximum''), and $\alpha$ values $\{0.1, 1, 2\}$. All active learning configurations outperform the random baseline with significance ($p < 0.05$\footnote{\label{sig_test}Significance was assessed using Welch's T-test, which is T-test variant that assumes unequal variances is the test distributions.}). The best configuration, uncertainty type =``sum'', similarity type=``maximum'', and $\alpha=0.1$, was used in active selection. This configuration and other hyperparameters of the Surrogate Classifier were tuned on $L_{Q}^D$ (for details, see Tables \ref{active_learning_tune} and \ref{hyperparam_surrogate} in the Appendix).

\textbf{Active learning performance:} After the first round of active learning, performance of the Surrogate Classifier was evaluated to confirm the effectiveness of the active learning framework. Model training included the sets: $L_Y\coloneqq \{\mbox{284 YVnotes samples}\}$ and $L_R\coloneqq \{$532 random MIMIC train samples$\}$. YVnotes was used to train the Surrogate Classifier to improve its accuracy and thereby obtain a better uncertainty score. $L_R$ was partitioned into $L_R^I\coloneqq \{\mbox{288 initial training samples}\}$ and $L_R^P\coloneqq \{\mbox{244 remaining samples, } L_R - L_R^I\}$. For the first round of active selection, an initial model, $M_I$, was trained on $\{L_R^I \cup L_Y\}$ and used to select 400 MIMIC samples, $L_A$. $L_R^P$ was withheld when training $M_I$ to validate the active learning approach. Hyperparameters were tuned on $L_D\coloneqq \{\mbox{188 MIMIC development samples}\}$ (parameter values in Table \ref{hyperparam_surrogate} of the Appendix). 

Figure \ref{surrogate_active_learning_perf} presents the performance of four cases on $L_E\coloneqq \{\mbox{376 MIMIC test samples}\}$. For \textit{MIMIC-only initial}, \textit{+random}, and \textit{+active}, 10 runs were performed to account for variance in model initialization. For \textit{MIMIC-only initial} and \textit{+random}, the training sets are fixed, as all data is used each run. For \textit{+active}, the training set varies because only a subset of $L_A$ is randomly selected each run, so sampling variance is introduced. The error bars in Figure \ref{surrogate_active_learning_perf} indicate the standard deviation of the F1 scores across runs.
Comparing \textit{MIMIC-only initial} to \textit{initial} demonstrates that including YVnotes improves performance. Adding active samples to the initial training set yields a statistically significant improvement over adding random samples ($p < 0.06$\textsuperscript{\ref{sig_test}}), demonstrating the effectiveness of the active learning framework on the surrogate task. 

\begin{figure}[h]
\centering
\begin{minipage}[t]{\dimexpr.5\textwidth-1em}
    \centering
    \frame{\includegraphics[]{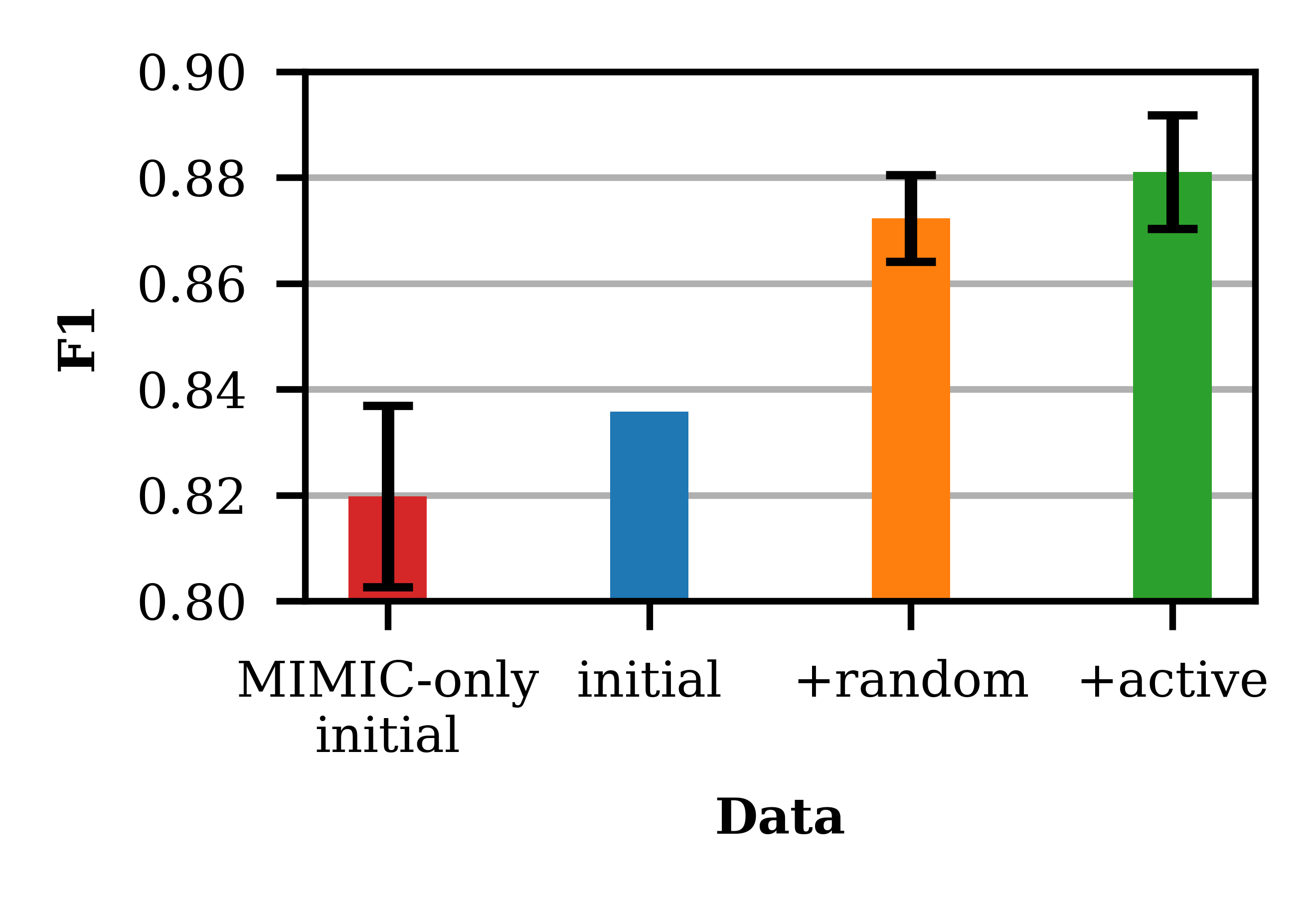}}
    \caption{Surrogate Classifier performance with random and active samples, evaluated on MIMIC test samples.}
    \label{surrogate_active_learning_perf}
\end{minipage}
\hfill
\begin{minipage}[t]{\dimexpr.5\textwidth-1em}
    \centering
    \frame{\includegraphics[]{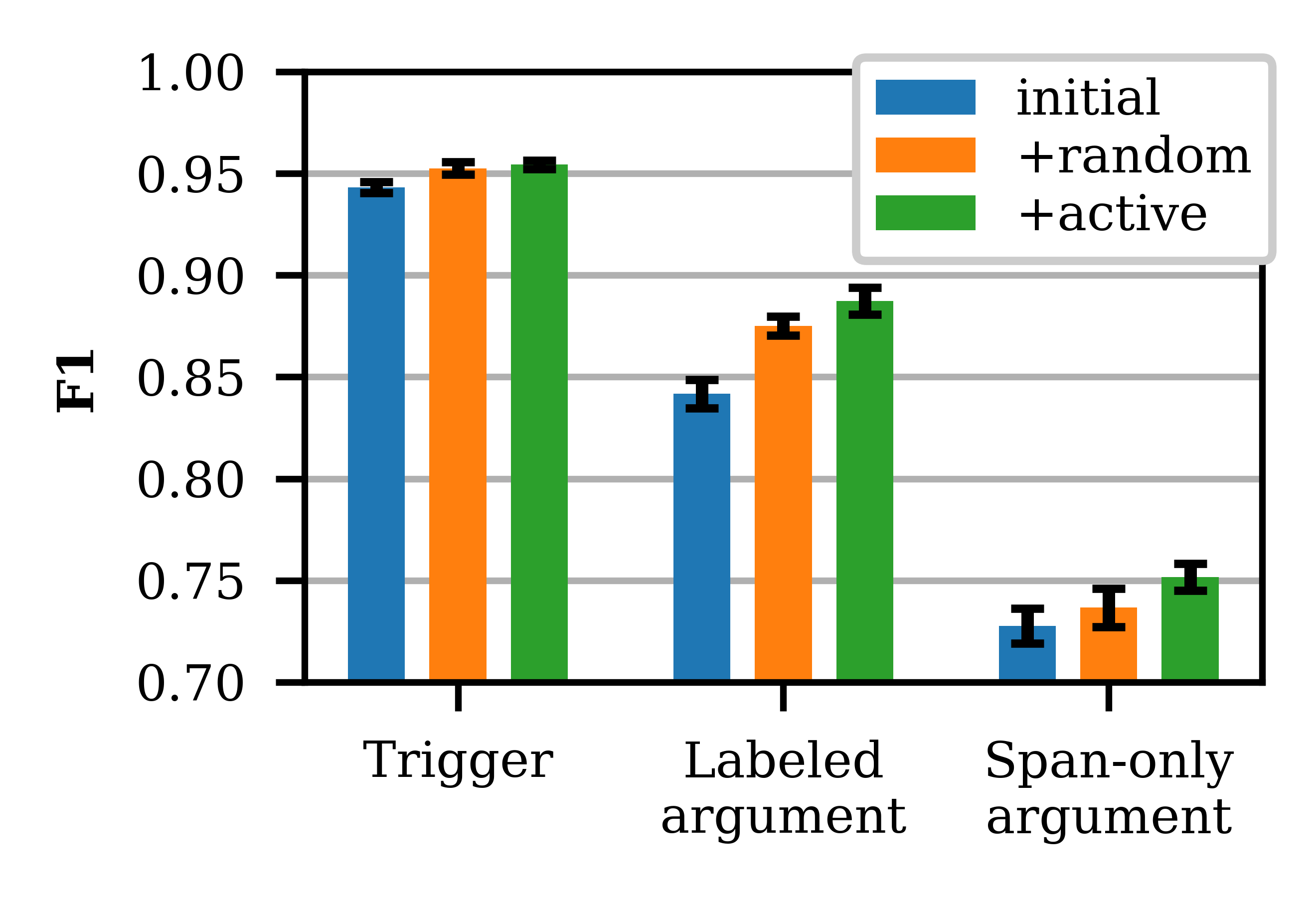}}
    \caption{Event Extractor performance with random and active samples, evaluated on MIMIC test samples.}
    \label{event_active_learning_perf}
\end{minipage}%
\end{figure}

The effectiveness of the active learning framework on the target event extraction task for the same conditions is presented in Figure~\ref{event_active_learning_perf}, where scores are averaged across event types.\footnote{For the Event Extractor, we exclude $L_Y$ since YVnotes do not include all of the labeled phenomena of SHAC.} The details of the event extraction model are presented in Section \ref{section_event_extraction}. The performance achieved by adding active samples outperforms that of adding random samples for labeled argument and span-only argument extraction, with significance ($p < 0.01$\textsuperscript{\ref{sig_test}}). The addition of actively selected notes improved extraction performance, relative to the random baseline, across most annotated phenomena. However, \textbf{the largest active learning performance gains were achieved for prominent health risk factors, including past and current drug use, current tobacco use, unemployment, homelessness, and living with others} (+0.09 $\Delta$F1 for \textit{current} \textit{Drug} \textit{Status}, +0.14 $\Delta$F1 for \textit{past} \textit{Drug} \textit{Status}, +0.07 $\Delta$F1 for \textit{current} \textit{Tobacco} \textit{Status}, +0.04 $\Delta$F1 for \textit{unemployed} \textit{Employment} \textit{Status}, +0.06 $\Delta$F1 for \textit{homeless} \textit{Living Status} \textit{Type}, and +0.07 $\Delta$F1 for \textit{with others} \textit{Living Status} \textit{Type}). The difference in trigger performance is not statistically significant. This result validates the use of the simplified surrogate text classification task as a proxy for the more complex event extraction task. After validating the active learning strategy, three additional rounds of active selection were performed (see Table \ref{annotation_rounds} of the Appendix for details), and the Surrogate Classifier model was retrained prior to each active round. Due to the limited number of random samples, further comparisons of active vs.\ random sampling are not possible.

\begin{figure}[h]
    \centering
    \frame{\includegraphics[]{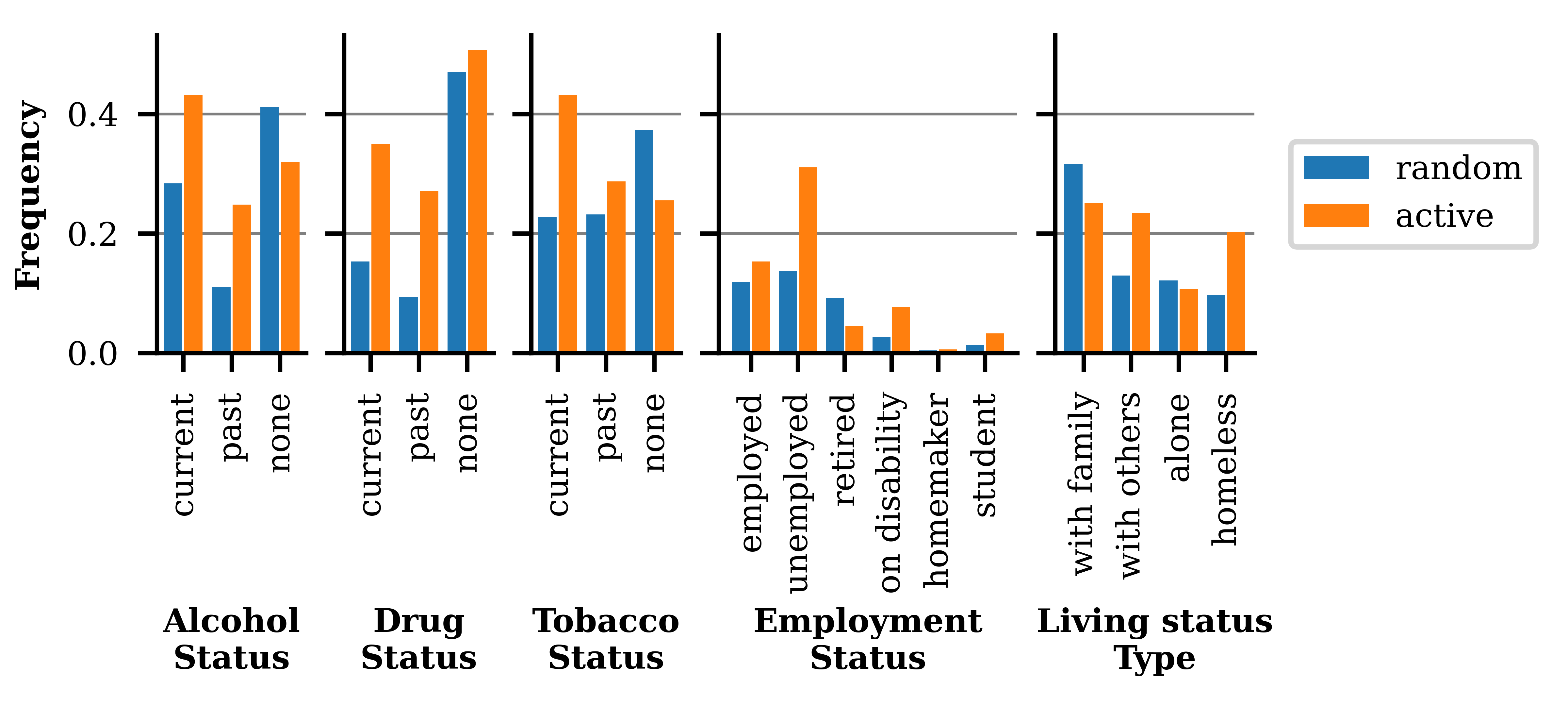}}
    \caption{Label frequency per social history section, comparing random and active sampling}
    \label{label_dist}
\end{figure}

We hypothesized the Surrogate Classifier uncertainty would bias the selection process to include more health risk factors (e.g. positive substance abuse, unemployment, being on disability, homelessness, etc.), which tend to be more challenging to automatically extract than less risky behavior (e.g. no substance use, being employed, and living with family). Active learning successfully identified samples with richer, more detailed SDOH descriptions. Figure \ref{label_dist} presents the label frequency per sample (note section) for random and active samples for the entirety of SHAC. The frequency of positive substance use ($\textit{Status} \in \{current, past\}$) is 83\% higher in active samples than random samples, with the frequency of positive drug use 151\% higher with active selection. Active sampling produced higher rates for all \textit{Employment} \textit{Status} labels, except \textit{retired}. Descriptions of retirement, tend to have low entropy, because of the reliable presence of keywords like ``retired" or ``retirement." Regarding \textit{Living Status}, the rate of \textit{homeless} is 109\% higher in active samples than random samples, and the rate of \textit{with others} is 81\% higher. The rate of \textit{alone} is slightly lower in active samples, likely due to lower entropy associated with the limited vocabulary used to describe living alone (e.g. ``alone'' or ``by herself'').

\section{Event Extraction}
\label{event_detection}

\label{section_event_extraction}
This section introduces the \textit{Event Extractor}, which jointly predicts all the phenomena in Table \ref{annotated_phenomena_partial}, and presents the initial extraction results for SHAC.

\subsection{Methods}
The Event Extractor generates sentence and token-level predictions that are assembled into events, similar to the SHAC annotation scheme. The Event Extractor builds on our previous state-of-the-art neural multi-task extractor for substance abuse information \citep{lybarger2018using}.
\begin{wrapfigure}{r}{0.5\textwidth}
    \centering
    \frame{\includegraphics[]{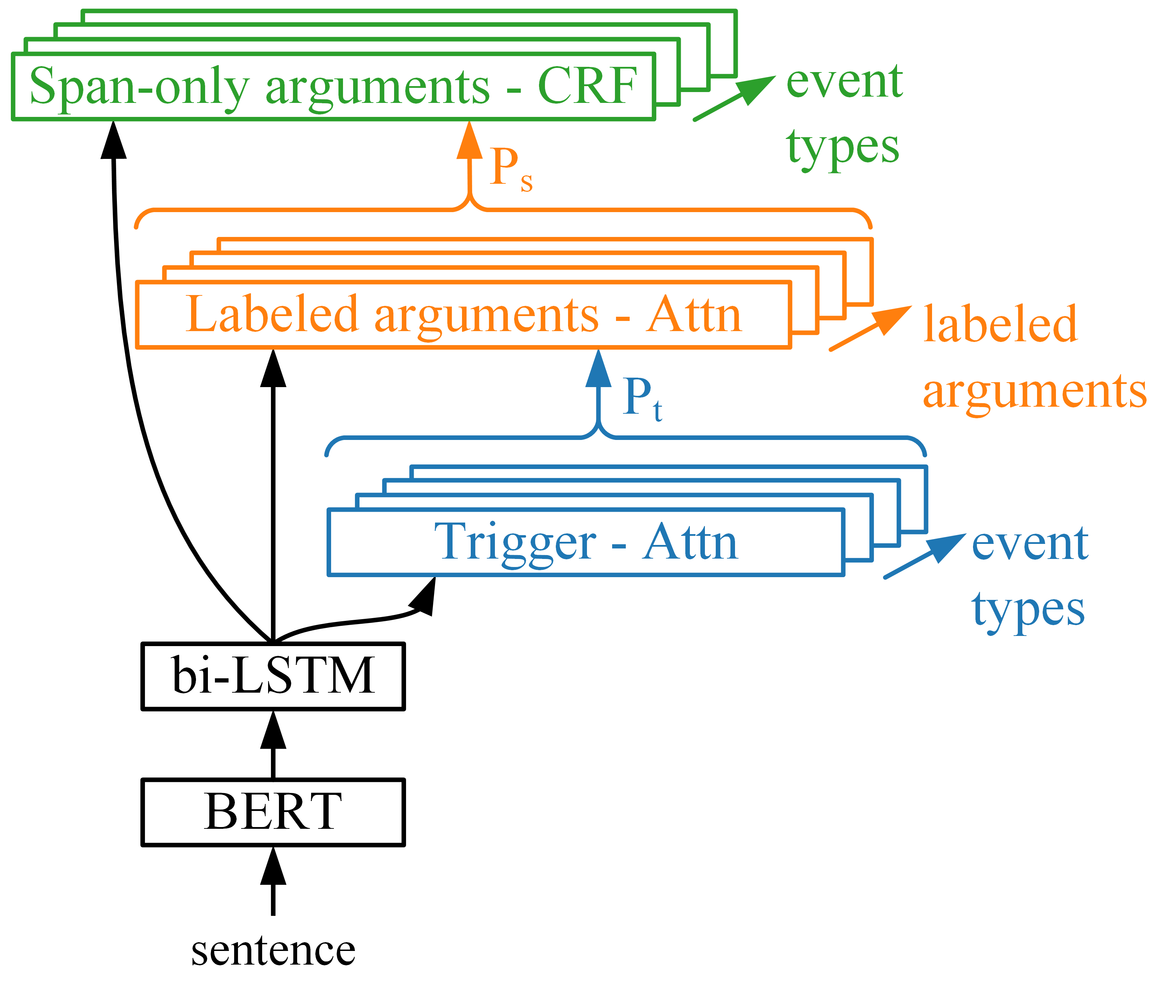}}
    
    \caption{Event Extractor model}
    \label{multitask_model}
\end{wrapfigure}
It is a generalized version of this previous work and is shown in Figure \ref{multitask_model}. 

\textbf{Shared layers:} Individual sentences are encoded using \textit{Bio+Discharge Summary BERT} \citep{LeeBioBERT2019}, creating an $n \times d$ matrix, where $n$ is the sentence length in tokens and $d$ is the BERT embedding size. BERT parameters are frozen during training (no back propagation) to limit computational cost. Similar to other work \citep{kitaev2018multilingual}, only the last word piece embedding for each token is used, to simplify the downstream sequence tagging. The BERT encoding feeds into a bi-LSTM. The forward and backward outputs states of the bi-LSTM are concatenated resulting in $n \times 2u$ matrix, $V$, where $u$ is the hidden size. $V$ feeds into event type and argument-specific output layers. 

\textbf{Trigger:} The presence of each event type is predicted using separate self-attentive binary classifiers (not present/present). Positive predictions serve as the trigger for assembling events, and the token position with the maximum attention weight serves as the trigger span. During training, event type $k$ is considered \textit{present}, if the sentence contains one or more events of type $k$. The trigger probability for event type $k \in \{1,..,m\}$ is calculated as
\begin{equation}
P_{k}^t= softmax(W_{k}^t (A_{k}^t V)^T+b_{k}^t)
\end{equation}
where 
$W_{k}^t$, is a $2 \times 2u$ weight matrix,
$b_{k}^t$ is a $2 \times 1$ bias vector, and $A_{k}^t$, is a $1 \times n$ vector of attention weights 
\begin{equation}
A_k^c=softmax(Y_k^c V^T) \mbox{ for } k=1,\ldots , m    
\label{attn}
\end{equation}
and $Y_k^c$ is another learned weight matrix.
The trigger probabilities, $P_k^t$, are concatenated to form a $2 \times m$ matrix, $P^t$, for the labeled argument prediction. An event is detected if it has probability greater than $50\%$.

\textbf{Labeled arguments:} Labeled argument prediction is also treated as a text-classification task, and utilizes separate self-attentive output layers for each labeled argument. The token position with the maximum attention weight serves as the argument span. The probability of labeled argument $l$ for event type $k$ is calculated as
\begin{equation}
P_{k,l}^s=softmax(W_{k,l}^s [P^t,(A_{k,l}^s V)^T] + b_{k,l}^s)
\end{equation}
where $W_{k,l}^s$ is a weight matrix, $A_{k,l}^s$ is a vector of attention weights, and $b_{k,l}^s$ is a bias vector. The dimension of $P_{k,l}^s$ depends on the number of possible labels for that event-argument combination. The labeled argument probabilities, $P_{k,l}^s$, are concatenated to form a $2 \times 6$ matrix, $P^s$, for use in span-only argument detection. Experimentation included six labeled arguments: \textit{Status} for \textit{Alcohol}, \textit{Drug}, and \textit{Tobacco}; \textit{Status} for \textit{Employment}; and \textit{Status} and \textit{Type} for \textit{Living status}.

\textbf{Span-only arguments:} Span-only arguments are predicted using linear-chain Conditional Random Field (CRF) \citep{Lafferty_2001_conditional} output layers at the output of the bi-LSTM, which is a popular sequence tagging approach \citep{lample2016neural, Luan_2017_scientific_info}. The bi-LSTM network learns sequential word dependencies, and the CRF learns conditional dependencies between labels. A separate CRF extracts the span-only arguments for each event type (i.e. five CRF output layers), with input features $V$ and $P^s$. Sequence labels are represented using the begin-inside-outside (BIO) approach. Experimentation included 20 span-only arguments: \textit{Duration}, \textit{History}, \textit{Type}, \textit{Amount}, and \textit{Frequency} for \textit{Alcohol}, \textit{Drug}, and \textit{Tobacco}; \textit{Duration}, \textit{History}, and \textit{Type} for \textit{Employment}; and \textit{Duration} and \textit{History} for \textit{Living status}.

\textbf{Training:} 
The Event Extractor was trained on the entire SHAC train set to simultaneously extract substance abuse, living situation, and employment information. Similar to previous multi-task work \citep{collobert2008unified, luo2017segment, jaques2016multi, liu2016attention, luan2018multi, harutyunyan2019multitask}, the Event Extractor shares information across tasks (event types and arguments in this application). The Event Extractor hyperparameters were tuned on the development set, $L_D$ (parameter values in Table \ref{hyperparam_surrogate} of the Appendix).

\subsection{Results}


\begin{wrapfigure}{r}{0.5\textwidth}
    \centering
    \frame{\includegraphics[]{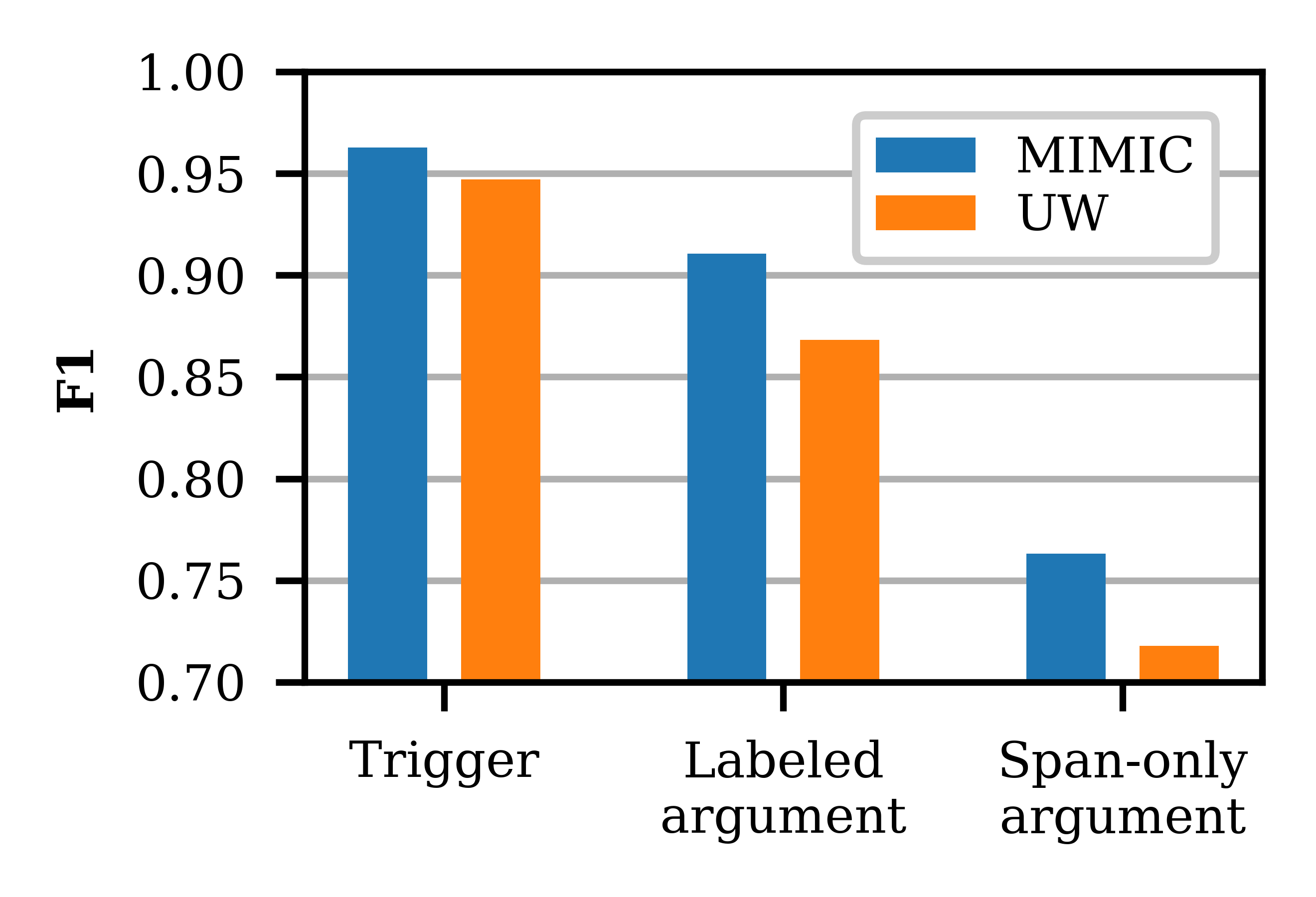}}
    \caption{Event Extractor average trigger and argument performance, comparing the MIMIC and UW Dataset test sets.}
    \label{event_extraction_perf}
\end{wrapfigure}

Figure \ref{event_extraction_perf} and Table \ref{event_extraction_perf_table} present the trigger and argument performance of the Event Extractor on the MIMIC and UW Dataset test sets. As described in Section \ref{annotator_agreement_scoring}, the argument extraction performance accounts for the alignment of the event triggers (i.e. only arguments with equivalent triggers can be equivalent). Overall, performance is higher on MIMIC, even though there are more UW Dataset training samples, including more active samples. The UW Dataset portion of SHAC includes four different note types, whereas the MIMIC portion includes only one note type, which likely contributes to the lower performance on the UW Dataset.  

\begin{table}[ht]
    \vspace{\wrapvspace}
    \small
    
\makeatletter
\def\hlinewd#1{%
  \noalign{\ifnum0=`}\fi\hrule \@height #1 \futurelet
   \reserved@a\@xhline}
\makeatother

\newcolumntype{L}[1]{>{\raggedright\arraybackslash}p{#1}}
\newcolumntype{C}[1]{>{\centering\arraybackslash}p{#1}}
\newcolumntype{R}[1]{>{\raggedleft\arraybackslash}p{#1}}

\begin{tabular}{|L{0.5in}|L{0.73in}|L{1.1in}|C{0.21in}|C{0.21in}|C{0.21in}|C{0.21in}|C{0.21in}|C{0.21in}|C{0.21in}|C{0.21in}|}
\hline
\multirow{2}{*}{\textbf{Field}}                                            & \multirow{2}{*}{\textbf{Event type}} & \multirow{2}{*}{\textbf{Argument}}                                                                        & \multicolumn{4}{c|}{\textbf{MIMIC}}                    & \multicolumn{4}{c|}{\textbf{UW}}                 \\ \cline{4-11} 
                                                                               &                                      &                                                                                                           & \textbf{\#} & \textbf{P} & \textbf{R} & \textbf{F1} & \textbf{\#} & \textbf{P} & \textbf{R} & \textbf{F1} \\ \hlinewd{1.0pt}
\multirow{5}{*}{Trigger}                                                       & Alcohol                              & --                                                                                                        & 314    & 0.99    & 0.96    & 0.97    & 404    & 0.97    & 0.99    & 0.98        \\ \cline{2-11} 
                                                                               & Drug                                 & --                                                                                                        & 194    & 0.96    & 0.95    & 0.96    & 481    & 0.97    & 0.92    & 0.94        \\ \cline{2-11} 
                                                                               & Tobacco                               & --                                                                                                        & 324    & 0.98    & 0.95    & 0.97    & 432    & 0.97    & 0.97    & 0.97        \\ \cline{2-11} 
                                                                               & Employment                           & --                                                                                                        & 169    & 0.93    & 0.96    & 0.94    & 148    & 0.86    & 0.91    & 0.89        \\ \cline{2-11} 
                                                                               & Living status                        & --                                                                                                        & 244    & 0.96    & 0.97    & 0.97    & 343    & 0.93    & 0.88    & 0.90         \\ \hlinewd{1.0pt}
\multirow{6}{*}{\begin{tabular}[c]{@{}l@{}}Labeled \\ argument\end{tabular}}   & Alcohol                              & Status                                                                                                    & 314    & 0.92    & 0.89    & 0.90    & 404    & 0.92    & 0.94    & 0.93          \\ \cline{2-11} 
                                                                               & Drug                                 & Status                                                                                                    & 194    & 0.91    & 0.89    & 0.90    & 481    & 0.85    & 0.80    & 0.82            \\ \cline{2-11} 
                                                                               & Tobacco                               & Status                                                                                                    & 324    & 0.91    & 0.89    & 0.90    & 432    & 0.91    & 0.90    & 0.90              \\ \cline{2-11} 
                                                                               & Employment                           & Status                                                                                                    & 169    & 0.84    & 0.88    & 0.86    & 148    & 0.79    & 0.83    & 0.81         \\ \cline{2-11} 
                                                                               & \multirow{2}{*}{Living status}       & Status                                                                                                    & 244    & 0.96    & 0.95    & 0.96    & 343    & 0.92    & 0.86    & 0.89        \\ \cline{3-11} 
                                                                               &                                      & Type                                                                                                      & 244    & 0.93    & 0.93    & 0.93    & 343    & 0.85    & 0.78    & 0.81         \\ \hlinewd{1.0pt}
\multirow{5}{*}{\begin{tabular}[c]{@{}l@{}}Span-only \\ argument\end{tabular}} & Alcohol                              & \multirow{3}{*}{\begin{tabular}[c]{@{}l@{}}Amount, Duration, \\ Frequency, History, \\ Type\end{tabular}} & 396    & 0.70    & 0.74    & 0.72    & 420    & 0.67    & 0.80    & 0.73            \\ \cline{2-2} \cline{4-11} 
                                                                               & Drug                                 &                                                                                                           & 219    & 0.67    & 0.75    & 0.71    & 583    & 0.62    & 0.63    & 0.62        \\ \cline{2-2} \cline{4-11} 
                                                                               & Tobacco                               &                                                                                                           & 799    & 0.81    & 0.83    & 0.82    & 880    & 0.78    & 0.81    & 0.79         \\ \cline{2-11} 
                                                                               & Employment                           & \begin{tabular}[c]{@{}l@{}}Duration, History, \\ Type\end{tabular}                                        & 441    & 0.80    & 0.74    & 0.77    & 261    & 0.77    & 0.77    & 0.77          \\ \cline{2-11} 
                                                                               & Living status                        & Duration, History                                                                                         &  21    & 0.21    & 0.57    & 0.31    &  57    & 0.19    & 0.26    & 0.22             \\ \hline
\end{tabular}

    \caption{Event Extractor trigger and argument role performance trained on the entire SHAC train set, evaluated on the MIMIC and UW Dataset test sets.}
    \label{event_extraction_perf_table}
    \vspace{\wrapvspace}
\end{table}
Table \ref{event_extraction_perf_table} presents detailed results for the same Event Extractor model and data configuration as Figure \ref{event_extraction_perf}. Trigger performance is greater than 0.89 F1 for all event types in both data sets. Labeled argument performance is similar in both data sets for \textit{Alcohol} and \textit{Tobacco} \textit{Status}; however, there are performance differences for \textit{Drug}, \textit{Employment}, and \textit{Living status} labeled arguments. In substance use \textit{Status} prediction, the \textit{none} label is typically less confusable and easier to predict than \textit{past} and \textit{current}. In the test set, the relative frequency of \textit{none} \textit{Status} labels for \textit{Drug} events is higher in MIMIC samples (80\%) than UW Dataset samples (57\%), which contributes to the higher performance on MIMIC. \textit{Living status} \textit{Status} performance is lower in the UW Dataset, even though the distribution of \textit{Status} labels is similar in both data sets. \textit{Living status} \textit{Type} performance is 0.12 F1 higher in MIMIC than the UW Dataset. In the test set, the distribution of \textit{Living status} \textit{Type} labels differs greatly between the data sets with the UW Dataset at 37\% \textit{with family}, 22\% \textit{with others}, 26\% \textit{homeless}, and 15\% \textit{alone} and MIMIC at 57\% \textit{with family}, 16\% \textit{with others}, 2\% \textit{homeless}, and 25\% \textit{alone}. For the span-only arguments, the performance is calculated at the token-level and micro averaged across the arguments for each event type. Span-only argument performance is comparable for \textit{Alcohol}, \textit{Tobacco}, and \textit{Employment}. However, it is higher for \textit{Drug} span-only arguments in MIMIC than the UW Dataset. \textit{Living status} span-only argument performance is very low for both data sets, primarily due to sparsity in the training set (only 167 \textit{Duration} and \textit{History} arguments among 3,267 \textit{Living status} events).

\subsection{Limitations}
\begin{figure}[ht]
    \begin{adjustbox}{varwidth=\textwidth,fbox,center, padding=0ex 0ex 0ex 0ex, margin=0ex 0ex 0ex 0ex}
    
        \begin{subfigure}[b]{\textwidth}
            \framebox{
                \begin{minipage}[t]{5.77in}
                    \includegraphics[scale=.20]{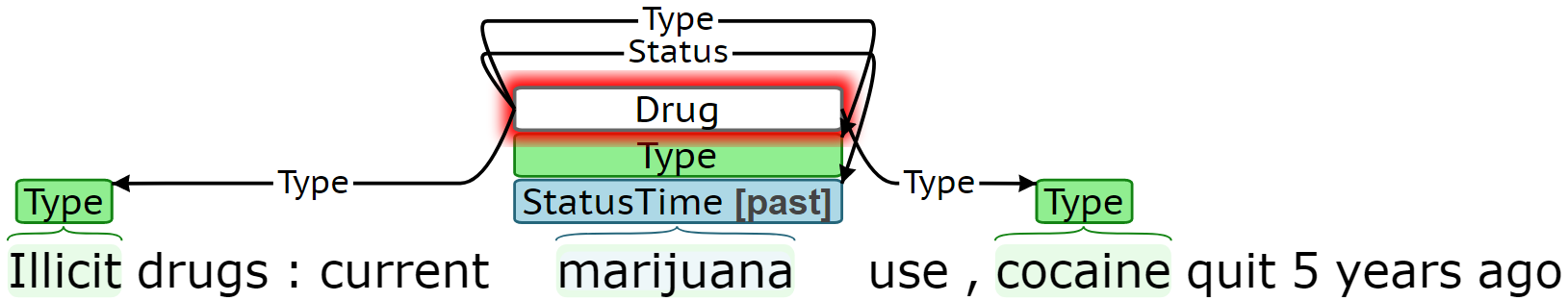}             
                \end{minipage}
            }
            \caption{Multiple gold drug events in one sentence}
            \label{error_analysis_multiple_drug}
        \end{subfigure}

        \begin{subfigure}[b]{\textwidth}
            \framebox{
                \begin{minipage}[t]{5.77in}
                    \includegraphics[scale=.20]{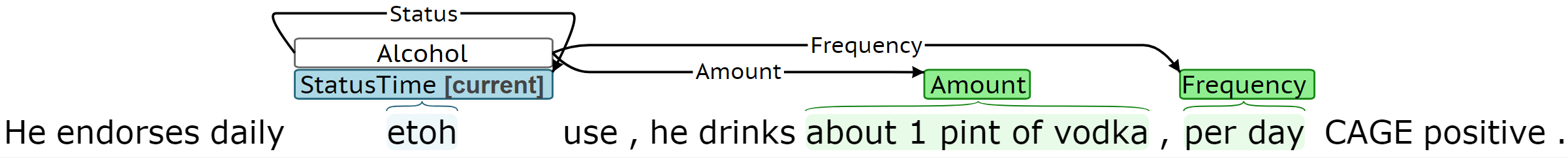}

                \end{minipage}
            }
            \caption{Gold span-only arguments overlap}
            \label{error_analysis_overlap}
        \end{subfigure}

        \begin{subfigure}[b]{\textwidth}
            \framebox{
                \begin{minipage}[t]{5.77in}
                    \includegraphics[scale=.20]{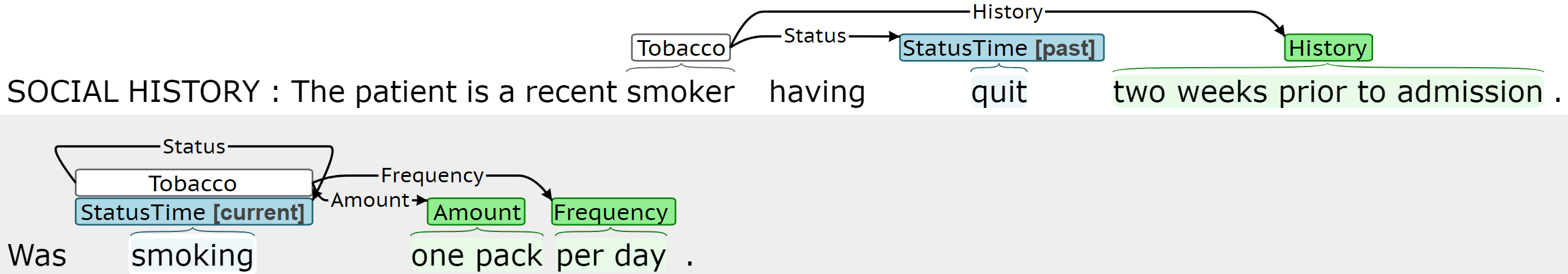}
                \end{minipage}
            }
            \caption{Intra-sentence information would likely benefit classifier}
            \label{error_analysis_intra_sentence}
        \end{subfigure}
    \end{adjustbox}
    \caption{Error analysis examples}
    \label{error_analysis}
\end{figure}

Although the Event Extractor achieved high performance for most target phenomena, the extraction framework has several limitations. The Event Extractor treats trigger and labeled argument prediction as a text classification task and can only represent a single event of a given type per sentence. Figure \ref{error_analysis_multiple_drug} presents predicted labels for a sentence with multiple gold \textit{Drug} events describing current marijuana use and previous cocaine use. While the \textit{Type} predictions in this example are correct, the \textit{Status} prediction of \textit{past} is incorrectly associated with both marijuana and cocaine. Of the sentences with at least one event in SHAC, 6\% contain multiple events of the same type. Span-only arguments for each event type are extracted using a single CRF, which cannot accommodate overlapping spans. Figure \ref{error_analysis_overlap} presents predictions for a sentence where the gold span-only argument spans overlap. The \textit{Amount} is correctly labeled as ``about 1 pint of vodka,'' but there should also be a \textit{Type} argument of ``vodka.'' Approximately 6\% of span-only arguments in events of the same type overlap in SHAC. The Event Extractor treats sentences independently. It does not incorporate context from the preceding sentences and cannot generate events that span multiple sentences. Figure \ref{error_analysis_intra_sentence} presents predictions for an example where past tobacco use is described in concurrent sentences. The first sentence includes a strong cue for \textit{past} \textit{Status}, ``quit''; however, the \textit{Status} in the second sentence is less clear without previous context. Fewer than 2\% of SHAC events span multiple sentences.

\section{Conclusions}

We present a new clinical corpus, SHAC, with detailed event-based annotations for 12 SDOH. SHAC includes approximately 4.5K social history sections from multiple institutions and note types and contains frequent descriptions of alcohol, drug, and tobacco use, employment, and living status. Approximately 71\% of the SHAC training set was selected using a novel active learning framework that utilizes a surrogate task for assessing sample uncertainty, which
increased the prevalence of critical risk factors in the annotated training data, including positive substance use, unemployment, disability, and homelessness, and increased event extraction performance, relative to using only randomly selected samples. The actively selected samples improve performance in both the surrogate task and the target event extraction task, validating the surrogate task approach. A neural multi-task model is presented for characterizing substance use, employment, and living status across multiple dimensions, including status, extent, and temporal fields. The event extractor model achieves high performance on the MIMIC and UW Dataset: 0.89-0.98 F1 in identifying distinct SDOH events, 0.82-0.93 F1 for substance use status, 0.81-0.86 F1 for employment status, and 0.81-0.93 F1 for living status type. The annotation guidelines and source code will be made available online\footnote{\url{https://github.com/uw-bionlp}}.

\section*{Acknowledgements}

This study was funded by the Seattle Flu Study through the Brotman Baty Institute and by the National Center For Advancing Translational Sciences of the National Institutes of Health under Award Number UL1 TR002319.

\bibliography{mybibfile}

\newpage
\section*{Appendix}

\setcounter{table}{0}
\renewcommand{\thetable}{A\arabic{table}}
\setcounter{figure}{0}
\renewcommand{\thefigure}{A\arabic{figure}}

\begin{table}[!htb]
\centering
\small

\begin{tabular}{|p{1.2in}|p{0.8in}|p{1.7in}|p{1.45in}|}
\hline
\textbf{Event type, $e$}                                   & \textbf{Argument type, $a$} & \textbf{Argument subtypes, $y_l$}                                                   & \textbf{Span examples} \\ \hline
\multirow{8}{1.2in}[\baselineskip]{Substance use (Alcohol, Drug, \& Tobacco)} & Status\textsuperscript{*}   & \{none, current, past\}                           & ``denies," ``smokes"                             \\ \cline{2-4} 
                                                            & Duration          & --                                                                  & ``for the past 8 years"                                            \\ \cline{2-4} 
                                                            & History           & --                                                                  & ``seven years ago"                                                 \\ \cline{2-4} 
                                                            & Type              & --                                                                  & ``beer," ``cocaine"                                         \\ \cline{2-4} 
                                                            & Amount            & --                                                                  & ``2 packs," ``3 drinks"                                           \\ \cline{2-4} 
                                                            & Frequency         & --                                                                  & ``daily," ``monthly"                                      \\ \hline 
\multirow{4}{1.2in}[\baselineskip]{Employment}              & Status\textsuperscript{*}            & \{\nohyphens{employed, unemployed, retired, \newline on disability, student, homemaker}\} & ``works," ``unemployed"               \\ \cline{2-4} 
                                                            & Duration          & --                                                                  & ``for five years"                                         \\ \cline{2-4} 
                                                            & History           & --                                                                  & ``15 years ago"                                                    \\ \cline{2-4} 
                                                            & Type              & --                                                                  & ``nurse," ``office work"                                 \\ \hline
\multirow{4}{1.2in}[\baselineskip]{Living status}           & Status\textsuperscript{*}            & \{current, past, future\}                            & ``lives," ``lived"                                         \\ \cline{2-4} 
                                                            & Type\textsuperscript{*}              & \{\nohyphens{alone, with family, with others, homeless}\}     & ``with husband"                         \\ \cline{2-4} 
                                                            & Duration          & --                                                                  & ``for the past 6 months"                                         \\ \cline{2-4} 
                                                            & History           & --                                                                  & ``until a month ago"                                               \\ \hline
Insurance                                                   & Status            & \{yes, no\}                                                             & ``has been off'''                       \\ \hline
\multirow{2}{1.2in}{Sexual orientation}                     & Status            & \{current, past\}                                                       & ``participated in''                       \\ \cline{2-4} 
                                                            & Type              & \{\nohyphens{heterosexual, homosexual, bisexual}\}                      & ``homosexual''                       \\ \hline
\multirow{2}{1.2in}{Gender identity}                        & Status            & \{current, past\}                                                       & ``identifies as''                       \\ \cline{2-4} 
                                                            & Type              & \{cisgender, transgender\}                                              & ``transgender''                       \\ \hline
Country of origin                                           & Type              & --                                                                  & ``England''                       \\ \hline
Race                                                        & Type              & --                                                                  & ``African American''                       \\ \hline
\multirow{6}{1.2in}{Physical activity}                      & Status            & \{none, current, past\}                                                 & ``currently jogs''                       \\ \cline{2-4} 
                                                            & Duration          & --                                                                  & ``for several years''                       \\ \cline{2-4} 
                                                            & History           & --                                                                  & ``10 years ago''                       \\ \cline{2-4} 
                                                            & Type              & --                                                                  & ``walks''                       \\ \cline{2-4} 
                                                            & Amount            & --                                                                  & ``4 miles''                       \\ \cline{2-4} 
                                                            & Frequency         & --                                                                  & ``every evening''                       \\ \hline
\multirow{6}{1.2in}{Environmental exposure}                 & Status            & \{none, current, past\}                                                 & ``no history''                       \\ \cline{2-4} 
                                                            & Duration          & --                                                                  & ``since 2001''                       \\ \cline{2-4} 
                                                            & History           & --                                                                  & ``until a month ago''                       \\ \cline{2-4} 
                                                            & Type              & --                                                                  & ``asbestos''                       \\ \cline{2-4} 
                                                            & Amount            & --                                                                  & ``significant''                       \\ \cline{2-4} 
                                                            & Frequency         & --                                                                  & ``daily''                       \\ \hline
\end{tabular}

\caption{Annotation guideline summary for all event types. *indicates the argument is required.}
\label{annotated_phenomena_all}
\end{table}

\begin{table}[!htb]
\centering
\small
\renewcommand{\arraystretch}{0.90}

\newcolumntype{L}[1]{>{\raggedright\arraybackslash}p{#1}}
\newcolumntype{C}[1]{>{\centering\arraybackslash}p{#1}}
\newcolumntype{R}[1]{>{\raggedleft\arraybackslash}p{#1}}

\begin{tabular}{|C{0.4in}|p{0.73in}|p{0.64in}|p{1.5in}|R{0.31in}|R{0.31in}|R{0.31in}|R{0.31in}|}
\hline
\textbf{Round} & \textbf{Source} & \textbf{Selection} & \textbf{Active learning training set}          & \textbf{Train} & \textbf{Dev} & \textbf{Test} & \textbf{Total} \\ \hline
1              & MIMIC           & Random                  & --                                               & 100            & --           & --            & 100            \\ \hline
2              & MIMIC           & Random                  & --                                               & 144            & 56           & --            & 200            \\ \hline
3              & MIMIC           & Random                  & --                                               & 288            & 112          & --            & 400            \\ \hline
4              & UW Dataset      & Random                  & --                                               & 84             & 140          & 280           & 504            \\ \hline
5              & MIMIC           & Active                  & 572 samples (Round 3 train + 284 YVnotes)        & 400            & --           & --            & 400            \\ \hline
6              & UW Dataset      & Random                  & --                                               & 168            & 120          & 240           & 528            \\ \hline
7              & MIMIC           & Random                  & --                                               & --             & 20           & 280           & 300            \\ \hline
8              & UW Dataset      & Random                  & --                                               & 112            & --           & --            & 112            \\ \hline
9              & UW Dataset      & Active                  & 1336 samples (Rounds 3-8 \textit{train} + 284 YVnotes)  & 728     & --           & --            & 728            \\ \hline
10             & UW Dataset      & Active                  & 2064 samples (Rounds 3-9 \textit{train} + 284 YVnotes)  & 728     & --           & --            & 728            \\ \hline
11             & MIMIC           & Active                  & 3036 samples (Rounds 1-10 \textit{train} + 284 YVnotes) & 384     & --           & --            & 384            \\ \hline
12             & MIMIC           & Random                  & --                                               & --             & --           & 96            & 96             \\ \hline
\multicolumn{4}{|r|}{\textbf{TOTAL}}                                                                          & \textbf{3136}  & \textbf{448} & \textbf{896}  & \textbf{4480}  \\ \hline
\end{tabular}
\caption{Annotation round summary, including selection type (random versus active) and training data used in active selection.}
\label{annotation_rounds}
\end{table}

\begin{table}[!htb]
    \small
    \centering
    \begin{tabular}{|p{0.72in}|p{0.6in}|p{0.2in}|p{0.35in}|}
\hline
\textbf{Uncertainty} & \textbf{Similarity} & \textbf{$\alpha$} &     \textbf{F1} \\ \hline
     loop &      average & 1.0 &  0.788* \\ \hline
     loop &      maximum & 0.1 &  0.776* \\ \hline
     sum &      average &  2.0 & 0.788* \\ \hline
     sum &      maximum &  0.1 & 0.794* \\ \hline

\end{tabular}

    \caption{Active learning query function tuning performance. *indicates statistical significance ($p < 0.05$) relative to a random baseline of 0.752 F1.}
    \label{active_learning_tune}
\end{table}

\begin{table}[!htb]
\centering
\small
\renewcommand{\arraystretch}{0.90}

\newcolumntype{L}[1]{>{\raggedright\arraybackslash}p{#1}}
\newcolumntype{C}[1]{>{\centering\arraybackslash}p{#1}}
\newcolumntype{R}[1]{>{\raggedleft\arraybackslash}p{#1}}

\begin{tabular}{|L{2.0in}|C{1.5in}|C{1.5in}|}
\hline
\textbf{Parameter}       & \textbf{Query function selection in Table \ref{active_learning_tune}} & \textbf{Active learning evaluation in Figure \ref{surrogate_active_learning_perf}} \\ \hline
batch size               & 20                                & 100                                 \\ \hline
learning rate            & 0.001                             & 0.005                               \\ \hline
maximum gradient L2 norm & 1.0                               & 1.0                                 \\ \hline
maximum length           & 200                               & 200                                 \\ \hline
number of epochs         & 500                               & 500                                 \\ \hline
LSTM hidden size         & 100                               & 100                                 \\ \hline
dropout, input to LSTM   & 0.7                               & 0.4                                 \\ \hline
dropout, output of LSTM  & 0.0                               & 0.4                                 \\ \hline
dropout, self-attention  & 0.7                               & 0.4                                 \\ \hline
\end{tabular}
\caption{Surrogate Classifier hyperparameters}
\label{hyperparam_surrogate}
\end{table}

\begin{table}[htb]
\centering
\small
\renewcommand{\arraystretch}{0.90}



\newcolumntype{L}[1]{>{\raggedright\arraybackslash}p{#1}}
\newcolumntype{C}[1]{>{\centering\arraybackslash}p{#1}}
\newcolumntype{R}[1]{>{\raggedleft\arraybackslash}p{#1}}

\begin{tabular}{|L{2.0in}|C{2.2in}|}
\hline
\textbf{Parameter}       & \textbf{Figure \ref{event_active_learning_perf}, Figure \ref{event_extraction_perf}, and Table \ref{event_extraction_perf_table}} \\ \hline
batch size               & 50                         \\ \hline
learning rate            & 0.005                      \\ \hline
maximum gradient L2 norm & 0.5                        \\ \hline
maximum length           & 30                         \\ \hline
number of epochs         & 250                        \\ \hline
LSTM hidden size         & 100                        \\ \hline
dropout, input to LSTM   & 0.6                        \\ \hline
dropout, output of LSTM  & 0.4                        \\ \hline
dropout, self-attention  & 0.4                        \\ \hline
\end{tabular}
\caption{Event Extractor hyperparameters}
\label{hyperparam_event}
\end{table}

\end{document}